\pdfoutput=1

\documentclass[11pt]{article}
\usepackage[]{acl}
\usepackage{mdframed}
\usepackage[colaction]{multicol}
\usepackage{ragged2e}
\usepackage{times}
\usepackage{latexsym}
\usepackage{comment}
\usepackage[T1]{fontenc}
\usepackage[colorinlistoftodos,prependcaption,textsize=tiny]{todonotes}
\usepackage[utf8]{inputenc}

\usepackage{microtype}
\usepackage{graphicx}
\usepackage{caption}
\usepackage{subcaption}
\usepackage{tabularx}
\usepackage{booktabs}
\usepackage{enumitem}

\usepackage{array}
\usepackage{tikz}

\newcolumntype{M}[1]{>{\centering\arraybackslash}m{#1}}

\newcommand\tikzmark[2]{%
\tikz[remember picture,overlay] 
\node[inner sep=0pt,outer sep=-0.9pt] (#1){#2};%
}

\newcommand\link[2]{%
\begin{tikzpicture}[remember picture, overlay, >=stealth, shorten >= 2pt]
  \draw[->] (#1.east) to  (#2.west);
\end{tikzpicture}%
}

\newenvironment{itemize*}%
  {\begin{itemize}[leftmargin=*,noitemsep,topsep=0pt]%
  \smaller
    \setlength{\itemsep}{0.9pt}%
    \setlength{\parskip}{0.9pt}%
    \setlength{\topsep}{0.9pt}}%
  {\end{itemize}}

\newenvironment{enumerate*}%
  {\begin{enumerate}%
    \setlength{\itemsep}{0.9pt}%
    \setlength{\parskip}{0.9pt}%
    \setlength{\topsep}{0.9pt}}%
  {\end{enumerate}}
\usepackage[nameinlink]{cleveref}

\newcommand*{\affaddr}[1]{#1} %
\newcommand*{\affmark}[1][*]{\textsuperscript{#1}}
\newcommand*{\email}[1]{\texttt{#1}}

\title{What Factors Should Paper-Reviewer Assignments Rely On?\\Community Perspectives on Issues and Ideals in Conference Peer-Review}

\author{
Terne Sasha Thorn Jakobsen\affmark[1], Anna Rogers\affmark[1,2]\\
\affaddr{\affmark[1] \normalsize Copenhagen Center for Social Data Science, University of Copenhagen} \\
\affaddr{\affmark[2] \normalsize RIKEN Center for Computational Science} \\
\affmark[1]{\normalsize \email{terne.thorn@sodas.ku.dk}}, \hspace{0.cm} 
\affmark[2]{\normalsize \email{arogers@sodas.ku.dk}}
}

\begin{document}
\maketitle
\begin{abstract}
Both scientific progress and individual researcher careers depend on the quality of peer review, which in turn depends on paper-reviewer matching. Surprisingly, this problem has been mostly approached as an automated recommendation problem rather than as a matter where different stakeholders (area chairs, reviewers, authors) have accumulated experience worth taking into account. We present the results of the first survey of the NLP community, identifying common issues and perspectives on what factors should be considered by paper-reviewer matching systems. This study contributes actionable recommendations for improving future NLP conferences, and desiderata for interpretable peer review assignments.
\end{abstract}

\section{Introduction}

Peer review is increasingly coming under criticism for its arbitrariness. Two NeurIPS experiments \cite{,Price_2014_NIPS_experiment,CortesLawrence_2021_Inconsistency_in_Conference_Peer_Review_Revisiting_2014_NeurIPS_Experiment,BeygelzimerDauphinEtAl_2021_NeurIPS_2021_Consistency_Experiment} have shown that the reviewers are good at identifying papers that are clearly bad, but the agreement on the ``good'' papers appears to be close to random. Among the likely reasons for that are cognitive and social biases of NLP reviewers \cite[see overview by][]{RogersAugenstein_2020_What_Can_We_Do_to_Improve_Peer_Review_in_NLP}, fundamental disagreements in such an interdisciplinary field as NLP, 
and acceptance rates that are kept low\footnote{\url{https://twitter.com/tomgoldsteincs/status/1388156022112624644}} irrespective of the ratio of high-quality submissions.

Such arbitrariness leads to understandable frustration on the part of the authors whose jobs and graduation depend on publications, and it also means lost time and opportunities \cite{AczelSzasziEtAl_2021_billion-dollar_donation_estimating_cost_of_researchers_time_spent_on_peer_review,GordonPoulin_2009_Cost_of_NSERC_Science_Grant_Peer_Review_System_Exceeds_Cost_of_Giving_Every_Qualified_Researcher_Baseline_Grant} for science overall. Reviews written by someone who does not have the requisite expertise, or does not even consider the given type of research as a contribution, it is a loss for all parties: the authors do not get the intellectual exchange that could improve their projects and ideas, and reviewers simply lose valuable time without learning something they could use. It is also a loss for the field overall: less popular topics could be systematically disadvantaged, leading to ossification of the field \cite{ChuEvans_2021_Slowed_canonical_progress_in_large_fields_of_science}.

\begin{figure}[t]
    \centering
    \includegraphics[width=0.85\linewidth]{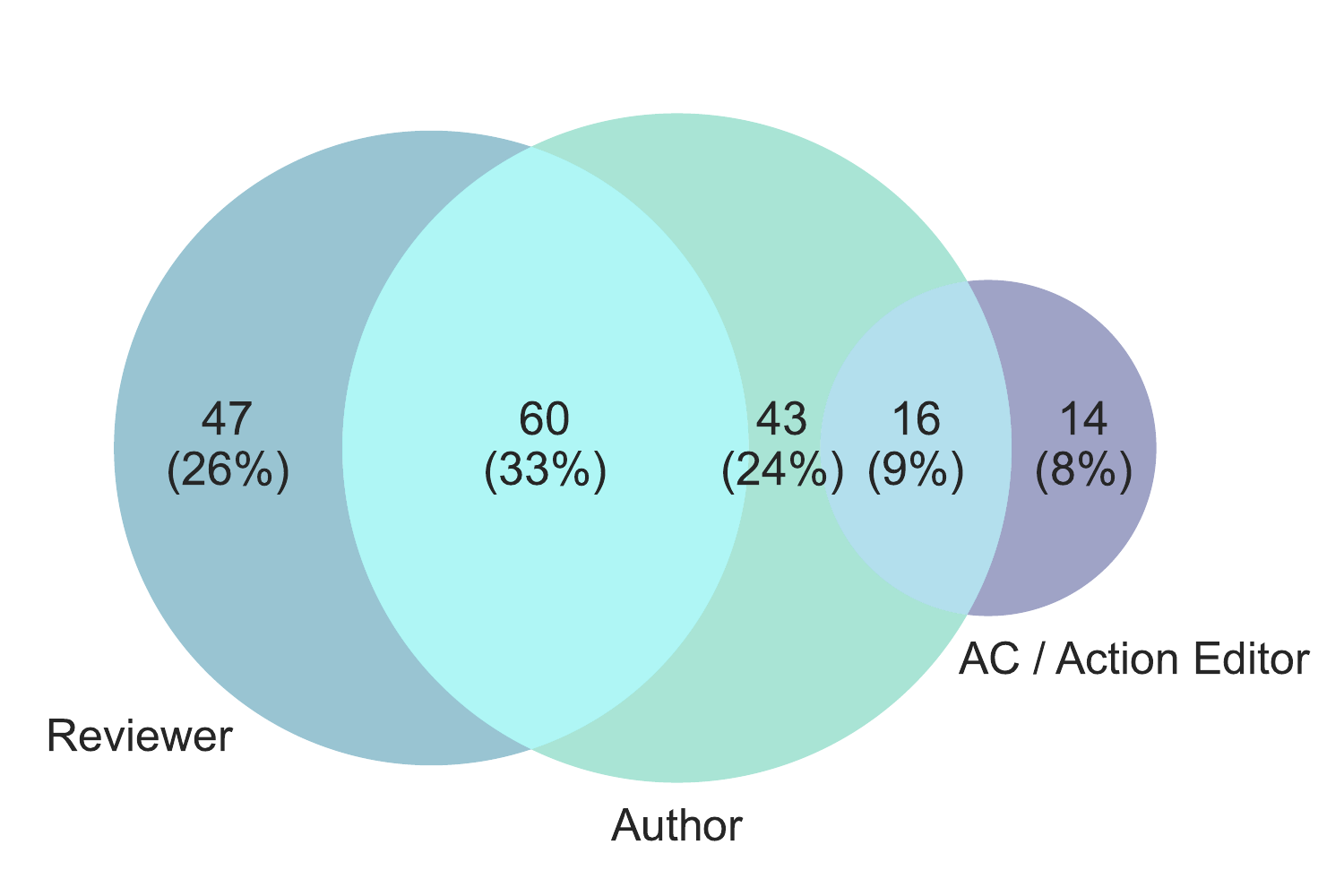}
    \caption{Overview of all respondents and overlap of their roles for their last experience at NLP venues.}
    \label{fig:venn}
\end{figure}

This paper contributes a snapshot of this problem in NLP %
venues, based on a survey of authors, reviewers and area chairs (ACs). We collected 180 responses, which is is comparable to the volume of feedback collected for implementing the ACL Rolling Review (ARR). The overall distribution of respondents' roles is shown in \cref{fig:venn}. We present the commonly reported issues and community preferences for different paper assignment workflows (\cref{sec:results}). We derive actionable recommendations to how peer review in NLP could be improved (\cref{sec:recommendations}), 
discuss the limitations of survey methodology (\cref{sec:limitation}), and conclude with desiderata for interpretable peer review assignments (\cref{sec:similarity}).

\section{Background: Peer Review in NLP}
\label{sec:background}
Paper-reviewer assignments are matches between submissions to conferences or journals and their available pool of reviewers, taking into account the potential conflicts of interest (COI) and reviewer assignment quotas. 

Among the systems used in recent NLP conferences, the Softconf matching algorithm takes into account bidding, quotas, and manual assignments, and randomly assigns the remaining papers as evenly as possible%
\footnote{\url{https://www.softconf.com/about/index.php/start/administration-view}}. NAACL and ACL 2021 used SoftConf, but also provided their ACs with affinity scores produced by a ``paraphrastic similarity'' system based on an LSTM encoder, which is trained on Semantic Scholar abstracts 
\cite{wieting-etal-2019-simple,2021_ACL_Reviewer_Matching_Code}. Affinity scores are scores indicating how well a given submission matches a given reviewer. They are typically computed as the similarity (e.g. cosine similarity) between the embeddings of certain information about the submission and the reviewer's publication history (e.g. abstracts and titles). 

ARR switched to OpenReview and currently uses\footnote{Source: personal communication with the ARR team.} their SPECTER-MFR system  \cite{2021_Paper-reviewer_affinity_modeling_for_OpenReview} which is based on SPECTER \cite{specter2020cohan} and MFR embeddings \cite{mfr}
 for computing affinity scores. The assignments are then made with the MinMax  matching algorithm\footnote{\url{https://github.com/openreview/openreview-matcher}}.

The problem of paper-reviewer assignment is by itself an active area of research (see overview of key issues for CS conferences by \citet{shah2022overview}). There are many proposals for paper-reviewer assignment systems \cite[][inter alia]{HartvigsenWeiEtAl_1999_Conference_Paper-Reviewer_Assignment_Problem,WangShiEtAl_2010_comprehensive_survey_of_reviewer_assignment_problem,LiWatanabe_2013_Automatic_Paper-to-reviewer_Assignment_based_on_Matching_Degree_of_Reviewers}, some of which also consider the problem of ``fair'' assignments %
\cite{LongWongEtAl_2013_On_Good_and_Fair_Paper-Reviewer_Assignment,StelmakhShahEtAl_2019_PeerReview4All_Fair_and_Accurate_Reviewer_Assignment_in_Peer_Review}.
Such studies tend to be hypothesis-driven: they make an assumption about what criteria should be taken into account, design a system and evaluate it. To the best of our knowledge, ours is the first study in the field to address the opposite question: what criteria should be taken into account, given the diversity of perspectives in an interdisciplinary field? We take that question to the community. 

\section{Methodology: survey structure and distribution}

We developed three separate surveys for the main groups of stakeholders in the peer review process: authors, reviewers and ACs.

They follow the same basic structure: consent to participation (see Impact Statement), background information, questions on most recent experiences in the role which the survey pertains to, and how the respondents believe paper-reviewer matching should be performed. 
Most questions are asked to respondents in all three roles, reformulated to match their different perspectives. 

The responses were collected late 2021 and all respondents are required to confirm that their most recent experience as an AC/reviewer/author is in 2019-2021. 
The full surveys and response data are publicly available\footnote{\url{https://github.com/terne/Paper-Reviewer-Matching-Surveys}}.

\paragraph{Participant background.}
All surveys include questions on career status and the number of times the respondents have been ACs/reviewers/authors at NLP venues. We ask what venues they have experience with (as broad categories) and what types of contributions they make in their work.

\paragraph{Participant experience with peer review.} We further ask the respondents a range of questions about their experience as AC/reviewer/author: 
how satisfied they are with the process, what issues they have experienced, what was the assignment load (ACs and reviewers), how paper-reviewer matching was done, %
how they would prefer it to be done, and which factors they believe to be important for paper-review matching. Most of the questions are multiple-choice, with addition of some open-ended 
questions where appropriate, so that respondents can elaborate their answers or add to the available options.  
Whenever possible, the question formulations were taken from the question bank of UK Data Service \cite{hyman2006use}. Attitude questions use a 5-point Likert scale.%

Limited memory is an important concern in surveys \cite{10.2307/2284504,Ayhan2005}, and we cannot expect the respondents to accurately recall all their experience 
with peer review. To reduce memory recall errors, the survey focuses on 
the respondent's most recent experience, but they also have a chance to reflect on prior experience in open-ended questions, 
and to report whether they experienced certain issues at any time in their career.

\paragraph{Survey distribution.} 
We distributed the surveys via three channels: by handing out flyers at EMNLP 2021, through mailing lists (ML-news, corpora list, Linguist list), and through Twitter with the hashtag \#NLProc. Participation was voluntary, with no incentives beyond potential utility of this study for improving NLP peer review.

\begin{figure}[t]
    \centering
    \includegraphics[width=\linewidth]{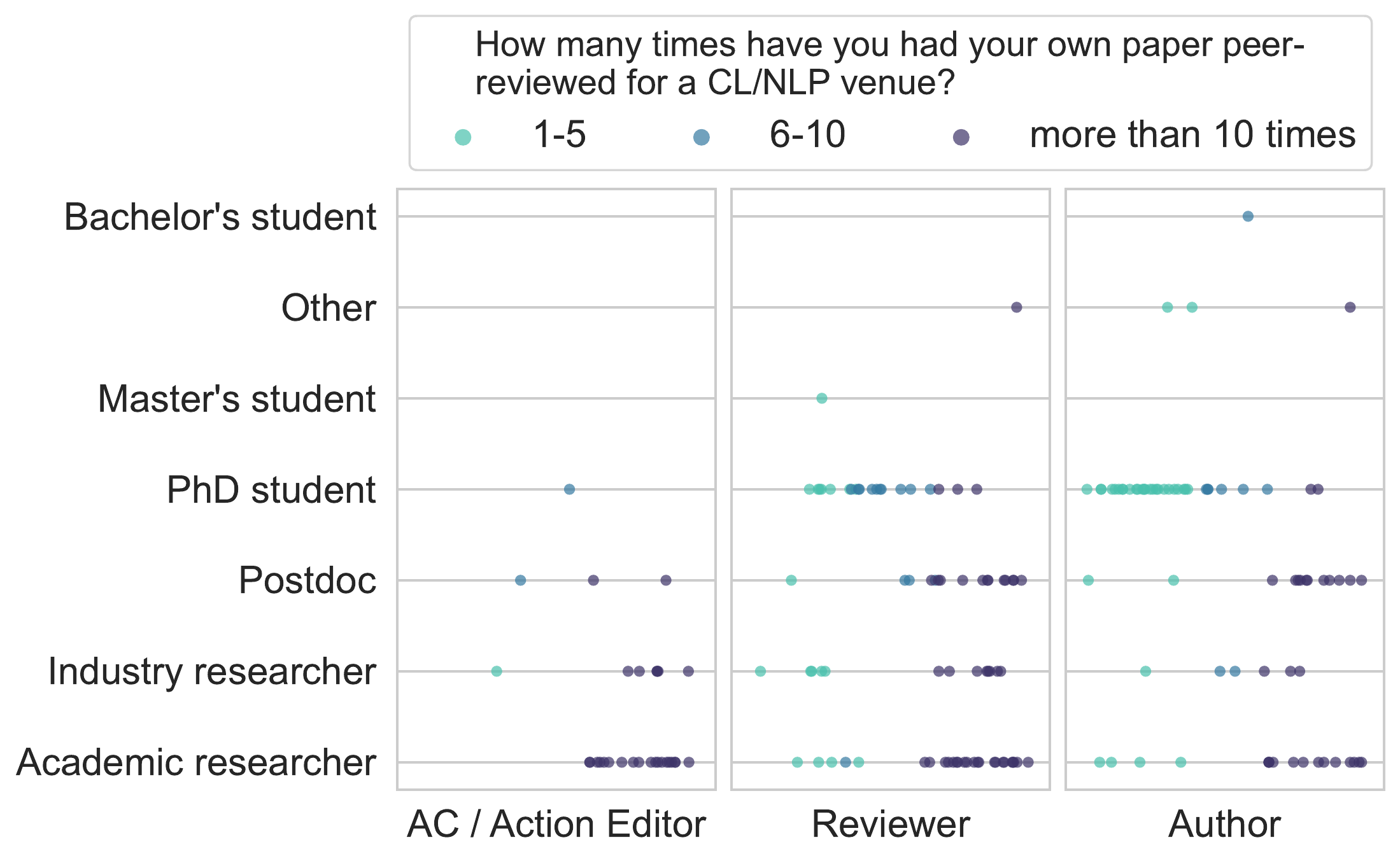}
    \caption{Career status of the respondents vs their experience receiving peer review. Numerical data is available in \cref{tab:background} in the appendix.}
    \label{fig:career}
\end{figure}

\paragraph{Data validation.}
Given that links to surveys were distributed openly and that we did not ask for any identifiable information, the surveys needed to include other means of validation to ensure that the responses included in the analysis were from attentive, relevant individuals. Our approach for validating the data quality follows \textit{satisficing theory} \cite{doi:10.1177/1470785317744856}, with the main safeguards being 1) the checking of response consistency, including a few ``traps" where inconsistency or illogical responses can be exposed, and 2) the inclusion of open-ended questions. 

73\% ACs, 40\% reviewers, 33\% authors have provided at least one response to our open-ended questions, and we did not find any meaningless or incoherent comments not addressing the question. For consistency checks, all respondents were asked: 

\begin{itemize}[leftmargin=*,noitemsep,topsep=0pt]
    \item How many times they have been an AC/reviewer/author. One of the options was ``0'', contradicting the earlier confirmation of experience in a given role.
    \item When was the last time they were an AC/reviewer/author. One of the options was ``earlier than 2019'', contradicting the earlier confirmation of peer review experience in 2019-2021.
    \item Whether they have performed the other roles. New authors may have not reviewed or AC-ed, but reviewers should also have been authors, and ACs should have experience with all roles.
\end{itemize}

\section{Results}
\label{sec:results}

Overall we received 38 responses from ACs, 87 from reviewers and 81 from authors (206 in total). After removing 20 incomplete responses and 8 responses inconsistent with the ``trap'' questions, we report the results for 30 
responses from ACs, 77 from reviewers and 73 from authors (180 in total).

\subsection{Who are the respondents?}\label{sec:who}

According to the past conference statistics, we could expect that many submissions would be primarily authored by the students, and reviewers are generally expected to be relatively senior, which should correspond to their going through peer review more often. We can use this expected pattern as an extra validation step for the survey responses.

\Cref{fig:career} shows that the responses are in line with this expected pattern. We received the most responses from academic researchers (62), PhD students (54), and postdocs (32). Most academic researchers and postdocs, but not PhD students, have had their work reviewed more than 10 times. At the same time 65\% of the PhD students who served as reviewers went through peer review more than 5 times, as opposed to 24.2\% of PhD students in the author role. Fewer industry than academic researchers responded to the survey. This could be related to the fact that a large part of the ``academic'' demographic are students -- and in 2020-2021 the ACL membership among students was equal to or exceeding other demographics \cite{acl2021-members}.

\begin{figure}[!b]
    \centering
    \includegraphics[width=.9\linewidth]{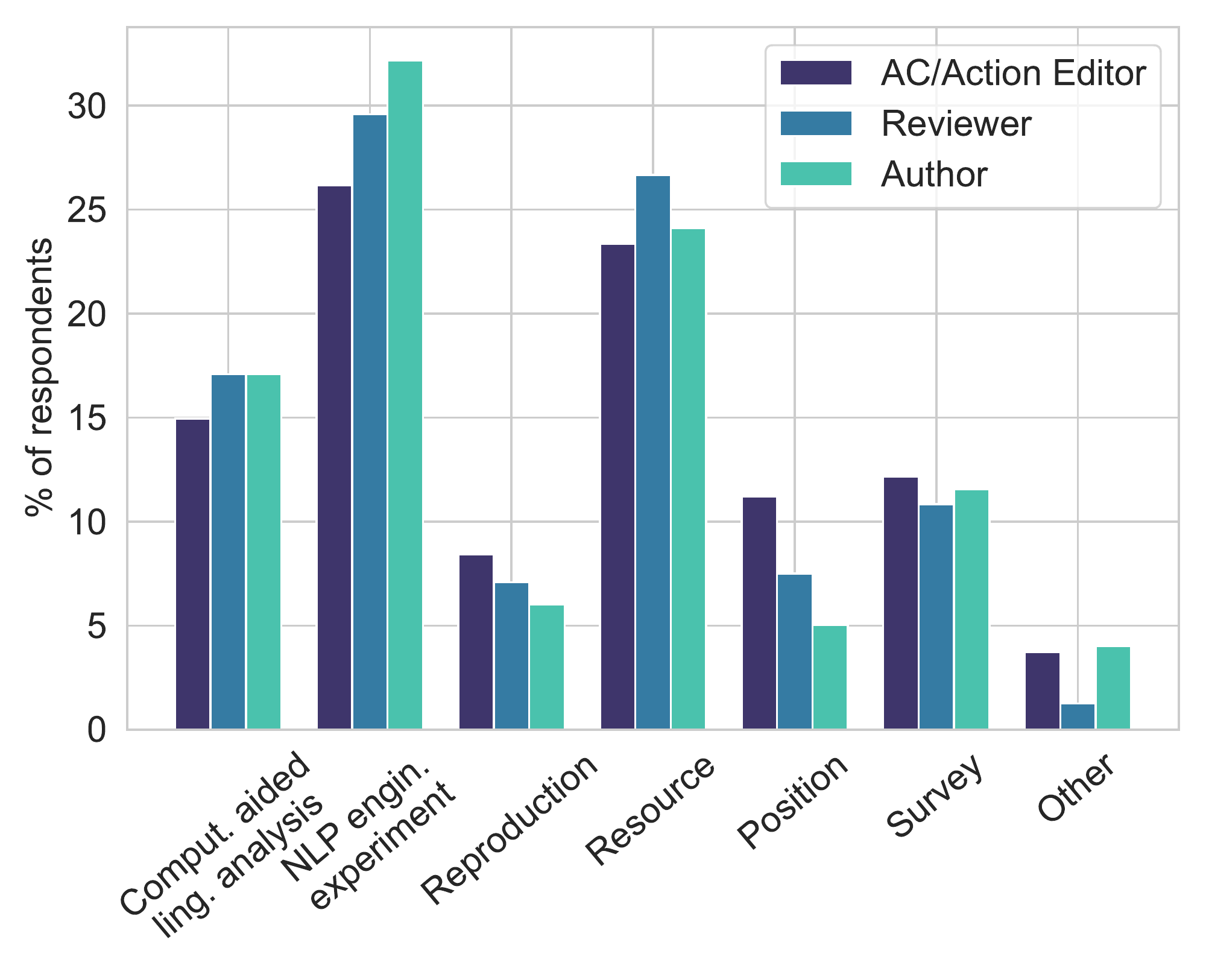}
    \caption{Types of research performed by respondents (multiple options could be selected).}
    \label{fig:papertypes}
\end{figure}

\subsection{Paper types}

The next question is to see what kinds of research papers the respondents to our surveys have authored: engineering experiment, survey, position paper etc., according to the COLING taxonomy by \citet{BenderDerczynski_2018_Paper_Types}%
. We expect that more senior researchers will have more experience with different types of work. 
Indeed, on average the authors have worked with 2.5 types of papers, vs. 3.0 for reviewers and 3.6 for ACs. The distribution is shown in \cref{fig:papertypes}. The most respondents have authored engineering experiment papers (with the authors reporting the most such work). 

Note that this only indicates whether the respondents to our surveys have or have not authored certain types of papers, rather than how many. In terms of volume, the engineering papers are a lot more prevalent: e.g. at ACL 2021 the ``Machine learning'' track had 332 submissions, vs 168 in the ``Resources and evaluation'' track \cite{XiaLiEtAl_2021_ACL-IJCNLP_2021_Program_Chair_Report}.

\subsection{What kinds of problems do people report?}

As with any voluntary feedback, our surveys were likely to receive more responses from people who had a grievance with the current process. Indeed, we find that only 6.7\% of ACs, 20.5\% of authors, and 22.1\% of reviewers say that they have not had any issues in their last encounter with NLP venues.

The overall distribution for the types of problems reported by the authors, reviewers and ACs in their last and overall experience 
is shown in \cref{fig:lastissues}. 
Given that at the time of this survey the ARR was recently deployed as the only ACL submission channel, we highlight the responses from the people for whom the most recent venue was ARR: 28\% %
reviewers, 18\% %
authors, 50\% ACs. %

The key takeaways are as follows:

\begin{itemize}[leftmargin=*,noitemsep,topsep=0pt]
\item Two of the most frequent complaints of ACs (about 50\% of the respondents) are insufficient information about reviewers and clunky interfaces;
\item Many paper-reviewer mismatches (about 30\%, if the report of the last experience is representative) are \textit{avoidable}: they should have been clear from the reviewers' publication history;
\item Over a third of the author respondents in their last submission (about 50\% over all history) received reviews from reviewers lacking either expertise or interest, and that is supported by the reviewers' reports of being assigned papers that were mismatched on one of these dimensions; 
\item The authors report that many reviews (over a third in last submission, close to 50\% overtime) are biased or shallow, which might be related to the above mismatches in expertise or interest;
\item Two patterns are exclusive to ARR: insufficient time for ACs\footnote{ARR has since switched to 6-week cycles, which might help to address this issue (\url{https://aclrollingreview.org/six-week-cycles/}).}, and zero authors with no issues.

\end{itemize}

\subsection{Knowledge of the workflow}

Our next question is what methods NLP venues use to match submissions to reviewers, and to what extent the stakeholders (authors and reviewers) are aware of how it is done. 
We find that relatively few authors (23.3\%) and reviewers (23.4\%) know for sure what process was used, which begs for \textit{more transparency in the conference process}. The ACs report that the most frequent case (37\%) is a combination of automated and manual assignments. Interestingly, most reviewers believe that their assignments were automated (36\%), and only (28\%) believe they were automated+manual. See App. \Cref{fig:whatis} for full distribution. 

\begin{figure}[t!]
         \includegraphics[width=0.95\linewidth]{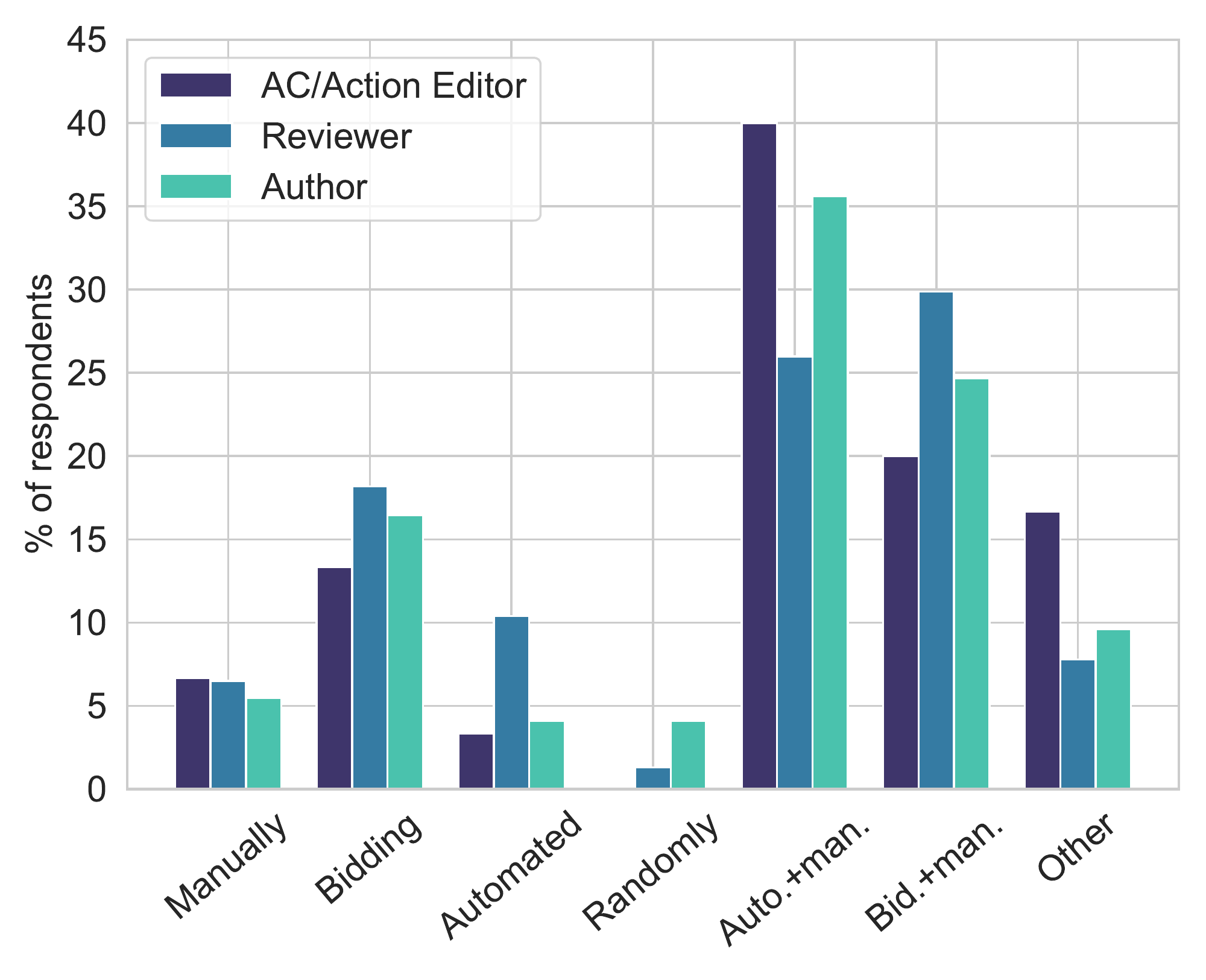}
         
\begin{subfigure}[t]{\linewidth}    
\begin{mdframed}
\scriptsize

\begin{center}\textbf{Topics mentioned in the open-ended comments}%

(See supplementary materials for full categorized comments)
\end{center}

\vspace{.5em}

\textbf{ACs:} bidding (2), similarity+manual (1), similarity+bidding+manual (5), keyword-based filtering + bidding (2), similarity (1), tracks (1), other info (2), ARR (1), interface (1)

\vspace{.5em}

\textbf{Reviewers:} manual (2), similarity + bidding (3), similarity+bidding+manual (3), keywords (1), keywords+similarity (1), tracks (2), tracks+bidding (1), other (4) 

\vspace{.5em}

\textbf{Authors:} against similarity (2), similarity + bidding (2), similarity+bidding+manual (2), ARR (2), random (2)
\end{mdframed}  
\end{subfigure}
\begin{minipage}{3.1cm}
\vfill
\end{minipage}
    \caption{%
    \textit{Which of the following options would you consider best for assigning reviewers to submissions?}%
    }
    \label{fig:whatshould}
\end{figure}

\section{The Ideal Process}
\label{sec:recommendations}

\subsection{Ideal workflow}

When asked about what paper-assignment process they would prefer (given that fully manual   

\onecolumn    
  \begin{figure}

    \centering
    \includegraphics[width=\textwidth]{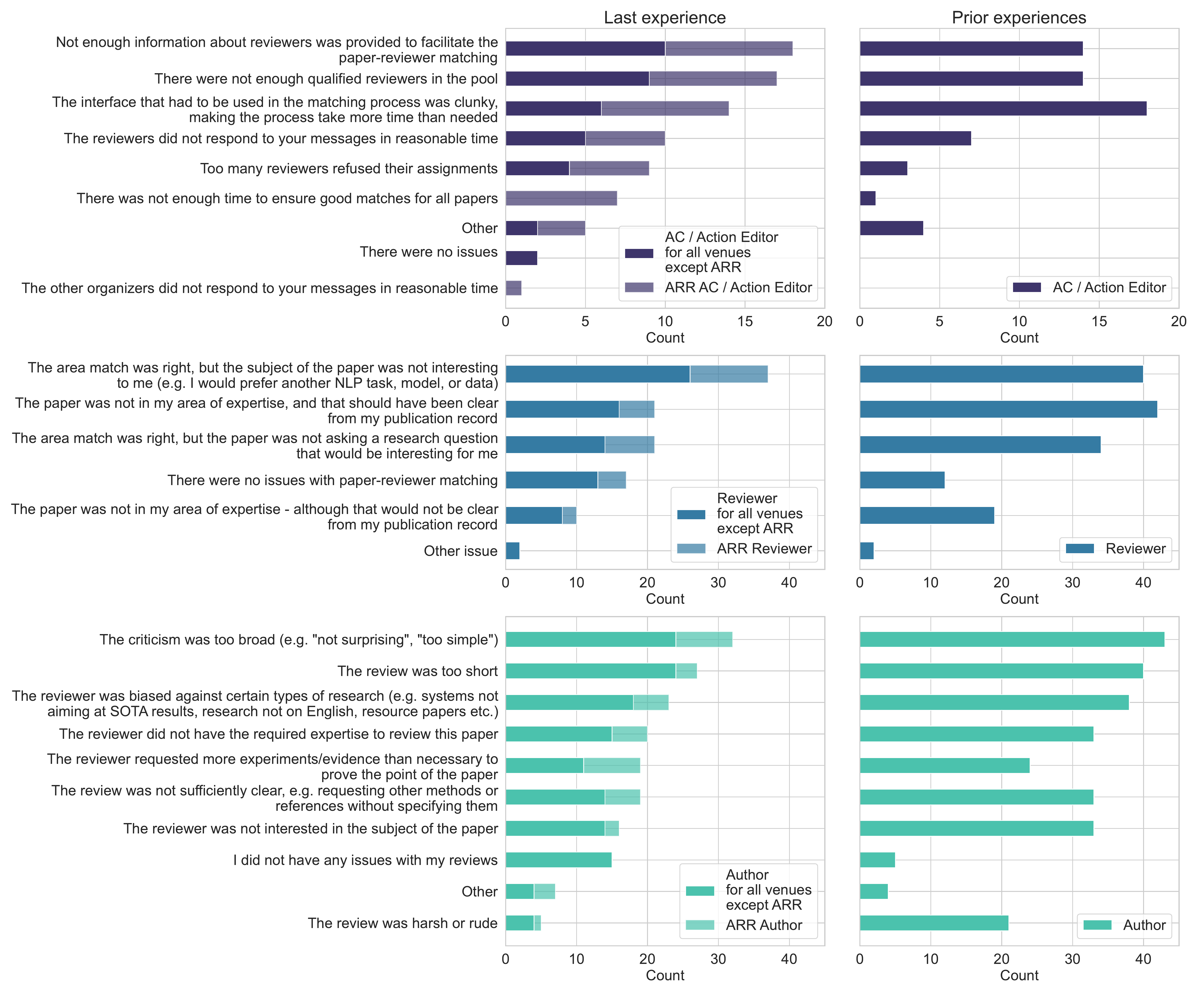}
\begin{mdframed}[userdefinedwidth=\textwidth]
\scriptsize

\begin{center}\textbf{Topics mentioned in the open-ended comments}

(The full comments categorized by these topics can be found in the survey data \href{https://github.com/terne/Paper-Reviewer-Matching-Surveys}{repository})
\end{center}

\vspace{1em}

\textbf{Area chairs:} interface issues (7), bad reviewers/reviews (5), workload issues (6), issues with ARR (4), lacking information on reviewers (4), communication issues between both systems and other human agents (4), lack of qualified reviewers in the pool (3), issues with meta-reviews (2), affinity score complaints (2), affinity score for finding reviewers the AC does not know personally (1), preference for manually recruited reviewers (1), papers assigned to ACs outside their area of expertise (1), too many declines (1), mismatch in goals of reviewers and authors (1), emergency reviews (1), bidding enabling bias (1).

\vspace{1em}

\textbf{Reviewers:} choices forced by ACs (5), preference for bidding (4), areas of past expertise not currently of interest (4), lack of interest in the paper (3), methodological mismatch between generations of NLP researchers (3), mismatch in research methods (2), publication records as an unreliable indicator for assignments (1), mismatch in languages (1), time issues (1), reviewer bias (1) 

\vspace{1em}

\textbf{Authors:} reviewer expectation for a certain kind of research (6), inattentive reviews (5), short reviews (3), mismatch between the score and the text of the review (3), requests for irrelevant citations (2), confirmation bias (1), non-constructive criticism (1), shallow reviews (1), lack of reviewer competence (2), missing reviews (2), requests for irrelevant comparisons (1), ``wild'' estimates of impact (1), unannounced policy changes (1)
\end{mdframed}            
    
    \caption{The issues with peer review process, reported by ACs, reviewers and authors, in their last (on the left) versus historical (on the right) experience with CL/NLP venues.}
    \label{fig:lastissues}    
\begin{multicols}{2}
\justifying
\noindent matching is impractical for large conferences), most ACs and authors opted for automated+manual process, but for the reviewers this is the second preferred process (26\%), with 30\% opting for bidding + manual checks (see \cref{fig:whatshould}). There was also relatively large support for pure bidding (13-18\% of respondents in all roles), and cumulatively pure bidding and bidding with manual adjustments have as much or more support from all respondent categories than the automated matching + manual assignments. 

The analysis of open-ended comments suggested that the respondents were aware that bidding is quite labor-intensive on the part of the reviewers. 5 ACs, 3 reviewers and 2 authors suggested using affinity scores to filter the papers on which bids would be requested, followed with manual checking. Another suggestion was keywords or more fine-grained areas/tracks, potentially as alternative to affinity scores for filtering down the list of papers to bid on. One AC suggested \textit{``an extensive, but still finite, set of tags (e.g. an ACL-version of}

\end{multicols}
\end{figure}
\twocolumn

\noindent\textit{ACM CCS concepts, or FAccT's submission tags''}. One reviewer stressed that the keywords should be provided by the authors, to match what \textit{they} perceive to be the salient aspects of the paper.

1 reviewer and 1 author suggested looking at whether the paper \textit{cites} the potential reviewer\footnote{We believe this is an interesting idea, but it could lead to authors strategically placing citations to maximize the chances of acceptance, or being punished for citing work  
that they may criticize or claim to improve upon.}, as this could be a good indicator for the reviewer's interest. 1 reviewer and 2 authors voiced support for some randomness in the assignments (given a track-level match): \textit{``Bidding + some random assignment to ensure diversity in the matching. We don't want reviewers to review only papers they *want* to review. However these random assignments should be clearly indicated to all, and treated accordingly.''}

\subsection{Ideal assignment criteria} \label{sec:ideal}

\textbf{AC past experience.} \Cref{fig:lastissues} shows that one of the most common problems for the ACs is that they were not provided with enough information to facilitate the paper-reviewer matching. The follow-up question is what information they \textit{are} provided with, and how useful they find it.

\begin{figure*}[t]
    \centering

    \includegraphics[width=\linewidth]{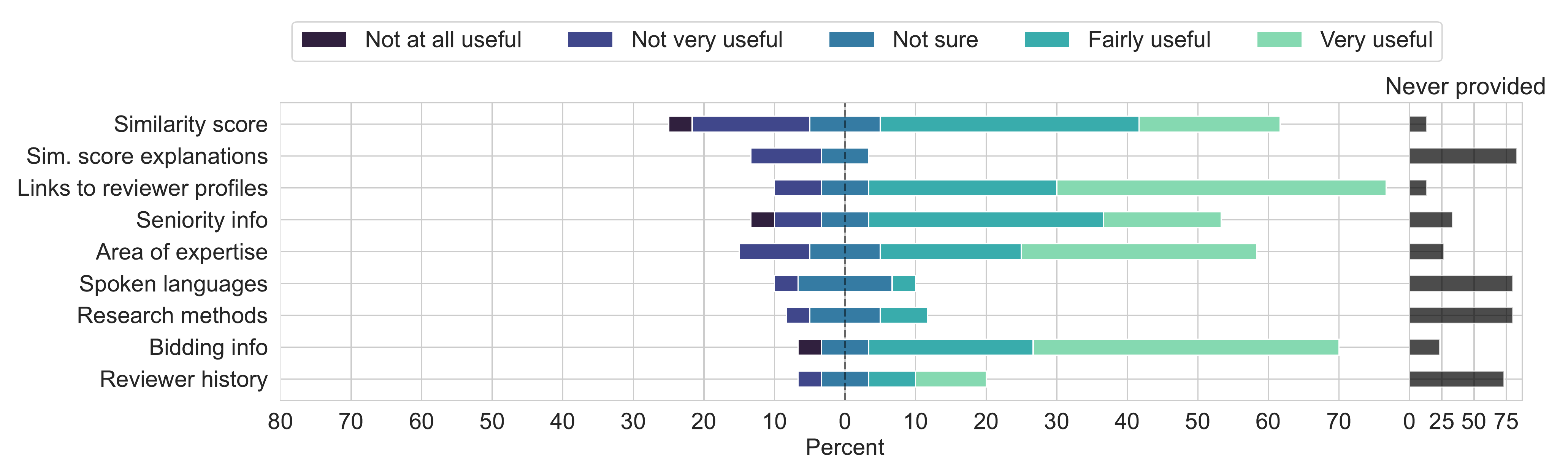}

\begin{mdframed}
\scriptsize
\textbf{Topics mentioned in the open-ended comments:} reviewer history (2), number of assigned papers (1), being able to ask SACs for advice (1), reviewer affiliation (e.g. academic or industry) (1), correct area match for both ACs and reviewers (1).
\end{mdframed}

    \caption{ 
    The diverging bars shows the experienced utility of different kinds of information about reviewers that ACs may have been presented with to assist in manual checks of paper-reviewer matches. If the respondent had never been presented with the specific kind of information they chose ``Never provided''.}   
\label{fig:usefulness}
\end{figure*}

\Cref{fig:usefulness} shows that the types of information with the highest utility information are links to reviewer profiles, bidding information, and affinity scores. 

\noindent But affinity scores are also the most controversial: it is the type of information that the most ACs find ``not very useful'' or ``not useful at all'' (20\%).

Overall the results suggest that ACs are presented with little structured information about reviewers, and have to identify the information they need from a glance at the reviewers' publication record. Seniority, expertise, and reviewer history notes from other ACs are all reported to be useful, but they were never provided directly to many ACs. 

An avenue for future research is offered by three types of information that the most ACs are not sure about, presumably because they are rarely provided: structured information about the methods that the reviewers were familiar with, the languages they spoke, and affinity score explanations. We will show below that there is much support for taking such methods into account. For the languages, this might be due to the ``default'' status of English \cite{Bender_2019_BenderRule_On_Naming_Languages_We_Study_and_Why_It_Matters}. We hypothesize that providing this information would make it easier to provide better matches for papers on other languages, which would in turn encourage the authors to submit more such work. Affinity will be discussed in \cref{sec:similarity}.

\begin{figure*}
    \centering

\includegraphics[width=\textwidth]{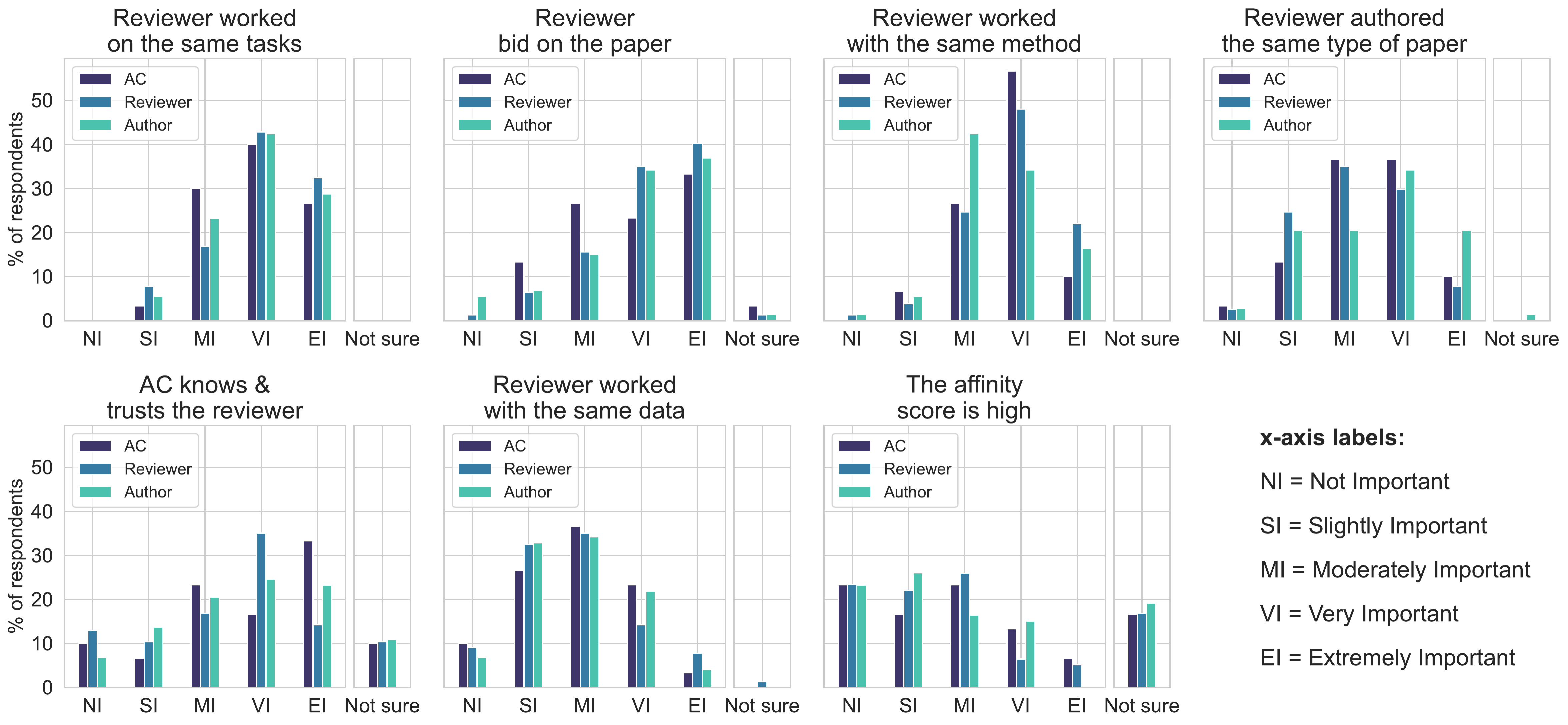}
    \caption{Question: \textit{How important do you think the following factors are for a good paper-reviewer match?}}
    \label{fig:importances}

\end{figure*}

\paragraph{Stakeholder preferences.} We then asked the respondents 
what factors they believe are important for paper-reviewer assignments. Their answers are shown in \cref{fig:importances}. The overall mean importance rankings (on scale 0-5) are as follows:

\begin{mdframed}[topline=false,rightline=false,leftline=false,bottomline=false,skipbelow=-0.5em,skipabove=-2.5em]
\footnotesize
\begin{tabular}{p{1pt}p{.3cm}p{6cm}}%
\tikzmark{a}{} & 3.95 & Reviewer has worked on the same task\\
 & 3.85 & Reviewer bid on the paper\\
 & 3.72 & Reviewer has worked with the same method \\
 & 3.32 & Reviewer has authored the same type of paper \\
 & 3.11 & AC knows \& trusts the reviewer \\
 & 2.81 & Reviewer has worked with the same kind of data \\
\tikzmark{c}{} & 1.99 & The affinity score is high \\
\end{tabular}
\link{a}{c}
\end{mdframed}

The fact that affinity scores rank the least important for NLP researchers (who would know the most about them) is interesting, and perhaps related to the fact that evaluation of paper-reviewer matching systems remains an open problem, with little empirical evidence for how well our current systems really work. In the absence of such evidence, our results suggest that the respondents across all groups are not very positive about their experience with such systems. In the authors' personal experience, when the conference chairs provide automated affinity scores they caution the area chairs against fully relying on them, and urge to adjust the assignments manually.

Our data suggests that within groups of stakeholders the individual variation in importance of different factors is higher for some factors and stakeholders than others: e.g. ACs vary within 1 point on the importance of knowing the data, but only within 0.74 points on importance of knowing the tasks. This has implications for approaches who would rely on AC assignments as ground truth for automated assignment systems: they could end up modeling the annotator instead of the task \cite{GevaGoldbergEtAl_2019_Are_We_Modeling_Task_or_Annotator_Investigation_of_Annotator_Bias_in_Natural_Language_Understanding_Datasetsa}. See App. \cref{tab:meanimportancescores} for full data.

We then explored the question of whether the experience of having authored research of a certain type correlates with any changes in the attitude towards some of these paper-reviewer matching factors. For each pair of type of research and matching factor, we ran two-sided Fisher's Exact tests for all respondents who have authored (or not) the types of research and the importance they attached to different factors in paper-reviewer assignment (binning on less than moderately important and more than moderately important). 
For some pairs there were statistically significant differences: e.g. the respondents who have authored reproduction papers were significantly more likely to believe it important that the reviewer has worked with the same kind of data ($p=0.004$), and respondents who authored position papers were significantly \textit{less} likely to believe a high automated affinity score is important ($p=0.003$). 
See \cref{tab:fisher1} in the appendix for all $p$-values and more details on the tests. We note that the relationships are not necessarily causal.

We conclude that our sample does provide evidence (the first, to our knowledge) that researchers in interdisciplinary fields who perform different kinds of research may have differing preferences for what information should be taken into account for paper-reviewer assignments. If that effect is robust, it should be considered in assignment systems for interdisciplinary fields. We hope that this finding would be explored in a larger study, taking into account both the experience of authoring a given type of paper and how central that type of research is for a given researcher (a factor that we did not consider). Another direction for future work is exploring this question from the perspective of demographic characteristics and the type of institution the respondents work in. Should there be significant differences, more targeted assignments could be a powerful tool for diversifying the field.

\subsection{Ideal workload}
\label{sec:workload}

We asked our reviewer and AC respondents how many assignments they received at their most recent NLP venue, and what would be the optimal number (given a month to review, and a week for AC assignments). For ACs, the mean optimal number of assignments is 8.5$\pm$4.2 vs. 9.1$\pm$5.1 they received at the most recent venue, and for reviewers it is 2.8$\pm$1.0 vs. 3.3$\pm$1.8. Whether this is an issue depends on how much time a given venue allows. The ARR reviewers have even less than a month, and they indicated preference for fewer assignments than they received (2.4$\pm$1.0 vs 3.3$\pm$1.9). 
See App. \cref{fig:revload} for data on other venues.

The lack of reviewers is a well-known problem. One of the possible causes is that many authors are students not yet ready to be reviewers. To investigate that, we asked the authors if they also reviewed for the venues where they last submitted a paper, and the reviewers and ACs - if they also submitted.  
If the core problem is that many authors are not qualified, we would expect more non-student authors to also be reviewers. Among all respondents there are 24\% authors who submit to a venue but do not review there or help in some other role (\cref{fig:venn}), but if we consider only non-student respondents that ratio is still 18\% (see non-student role distribution in App. \cref{fig:vennSenior}). This suggests that \textit{many qualified people do not review}.

\section{Discussion}

\subsection{Reviewer interests}

Our results suggest the lack of interest is one of the most common problems in paper-reviewer matching, for both authors and reviewers. The authors are aware of this problem and sometimes try to optimize for it by pursuing the ``safe'', popular topics. Unenthusiastic reviewers will likely produce shallow, heuristic-based reviews, essentially penalizing non-mainstream research. Both tendencies contribute to ossification of the field \cite{ChuEvans_2021_Slowed_canonical_progress_in_large_fields_of_science}, and generally need to be minimized.

It is in the AC's interest to find interested reviewers, since that minimizes late reviews, but they need to know who finds what interesting. That is not as simple as a match by topic/methodology, clear from the publication record. Interests change not only gradually over time but also according to what is popular or \textit{salient} at the given moment \cite{10.2307/1738360,DaiJane2020Psif}, or even in seemingly irrational ways (e.g. by being sensitive to the framing of the problem) \cite{TverskyA1981TFoD}. But although experience and knowledge may provide more stable descriptions of a reviewer, looking into dated publication records may be fundamentally counter-productive. According to one of our respondents: \textit{``I prefer the conferences who offer bidding processes to select the papers to review... I am more enthusiastic to review the papers compared to conferences that assign papers based on what my interests were x years ago.''}

Bidding however has its own set of problems, including the practical impossibility to elicit all preferences over a big set of papers, the possibility of collusion rings \cite{Littman_2021_Collusion_rings_threaten_integrity_of_computer_science_research}, and, as one of our respondents put it, ``\textit{biases towards/against certain paper types when bidding is enabled}''. But these problems potentially have solutions: there is work on detecting collusion rings \cite{BoehmerBredereckEtAl_2022_Combating_Collusion_Rings_is_Hard_but_Possible}, and several respondents suggested that bidding could be facilitated by subsampling with either keyword- or affinity-score-based approaches. 

We support some of our respondents' recommendation for a combination of interest-based and non-interest-based (within a matching area) assignments, with the latter clearly marked as such for ACs and reviewers, and separate playbooks for the two cases. The reviewer training programs should aim to develop the expectation that peer review is something that combines utility and exploration.

\subsection{Limitations}
\label{sec:limitation}

We readily acknowledge that, like with any surveys with voluntary participation, our sample of respondents may not be representative of the field overall, since the people who have had issues with peer review system are more incentivized to respond. However, precisely for that reason this methodology can be used to learn about the commonly reported types of problems, which was our goal. Our response rate turned out to be comparable to the response rate of the official ACL survey soliciting feedback on its peer review reform proposal \cite{Neubig_2020_ACL_Rolling_Review_Proposal},  %
which received 199 responses.

It is an open problem how future conferences could systematically improve, if they cannot rely on surveys to at least reliably estimate at what scale an issue occurs. Asking about satisfaction with reviews does not seem to produce reliable results \cite{Some_NAACL_2013_statistics_on_author_response_review_quality_etc,ACL_2018_Report_on_Review_Process_of_ACL_2018}. Our survey included a question about satisfaction with the paper-reviewer matching, and whether the most recent experience was better or worse than on average. Both reviewers and authors were more satisfied than dissatisfied, and considered the recent experience better than on average, despite reporting so many issues (see App. \cref{fig:satisfaction} for the distribution).

\subsection{Interpretable Paper-Reviewer Matching: Problem Formulation}
\label{sec:similarity}

There already are many proposed solutions for paper-reviewer matching (see \cref{sec:background}), but their evaluation is the more difficult problem. The obvious approach would be to use bidding information or real assignments made by ACs as ground truth, but this data is typically not shared to protect reviewer anonymity.  
It would also provide a very noisy signal not just due to different assignment strategies between ACs, but also different quality of assignments depending on how much time they have on a given day. Both ACs and bidding reviewers are also likely\footnote{Position bias is well documented in search \& recommendation systems \cite{CraswellZoeterEtAl_2008_experimental_comparison_of_click_position-bias_models,CollinsTkaczykEtAl_2018_Study_of_Position_Bias_in_Digital_Library_Recommender_Systems}.} to favor top-listed candidates. And, as our findings suggest, the optimal assignment strategies in an interdisciplinary field might genuinely vary between different types of papers and tracks. A system unaware of that might systematically disadvantage whole research agendas.

Given that even the human experts cannot tell what the best possible assignments are, we propose to reformulate the problem as \textit{interpretable paper-reviewer matching}. 
That problem is \textit{not} the same as the problem of faithfully explaining why a given paper-reviewer matching system produced a certain score, for which we have numerous interpretability techniques \cite{Sogaard_2021_Explainable_Natural_Language_Processing}. The AC goal is fundamentally different: not to understand the system, but to quickly find the information that the AC\footnote{Or the program chairs, should the conference aim to have consistent policies for all ACs.} considers relevant for making the best possible match. Therefore \textit{the task of interpretable paper-reviewer matching is rather to help to identify the information that the stakeholders wish the decisions to be based on, and to provide that information as justification for the decisions}.

\section{Conclusion}

We present the results of the first survey on paper-reviewer assignment from the perspective of three groups of stakeholders in the NLP community: authors, reviewers, and ACs. The results point at a host of issues, some immediately actionable (e.g. providing the ACs with better information), some normative (e.g. different kinds of research may need different assignment strategies), and some open (e.g. how do we evaluate the effect of any changes to peer review process?) A big issue for both authors and reviewers is mismatches due to lack of interest, which is in tension with explorative aspects of peer review. We recommend to address this issue with a combination of assignments based on bidding and random matches within area, backed up by reviewer training.

\section*{Acknowledgments}
Many thanks to Marzena Karpinska, Friedolin Merhout, and the anonymous reviewers for their insightful comments. We would also like to thank all our survey respondents, without whom this study would not have been possible.

\section*{Impact Statement}\label{sec:impact}

\paragraph{Broader impact.} The study identifies types of information that could be used to provide better paper-reviewer matches. Used strategically by a conference, it could be a powerful tool for diversifying the field, by helping the non-mainstream papers find the reviewers more open to them. By the same token, if the entity organizing the review process aimed for suppressing such research, de-prioritising this information could harm such papers. Our proposal of interpretable paper-reviewer assignments would mitigate this potential risk by requiring the organizers to disclose their rationale for any given match.

\paragraph{Personal data.} The surveys are designed to not solicit any personally identifiable information (including comments about individual peer review cases in the past conferences), or demographic information about participants.

\paragraph{Potential risks.} The respondents are participants in anonymous peer review process, and as such being tracked back to individual peer review cases could expose them to retaliation. The survey therefore did not solicit information about specific venues (only broader categories such as ``*ACL conferences''), and we manually verified that the open-ended comments also do not contain references to specific cases. We thus foresee no potential risks from deanonymization of the respondents.

\paragraph{Informed consent.} The respondents are informed about the organizers and the objective of the study: to identity current practises of paper-reviewer assignments in CL/NLP conferences and ways in which this process can be improved. Responses are anonymous and respondents consent to the use and sharing of their responses for research purposes. Respondents must give consent to continue the survey.

\paragraph{Intended use.} The survey data and forms will be made publicly available for research purposes.

\paragraph{Institutional approval.} The study was approved by the Research Ethics Committee at the authors' institution.

\bibliography{anthology,custom}
\bibliographystyle{acl_natbib}

\cleardoublepage

\appendix

\section{Appendix}
\label{sec:appendix}

In this appendix we introduce supplementary figures and tables.

\begin{figure}[h!]%
     \centering
     \includegraphics[width=\linewidth]{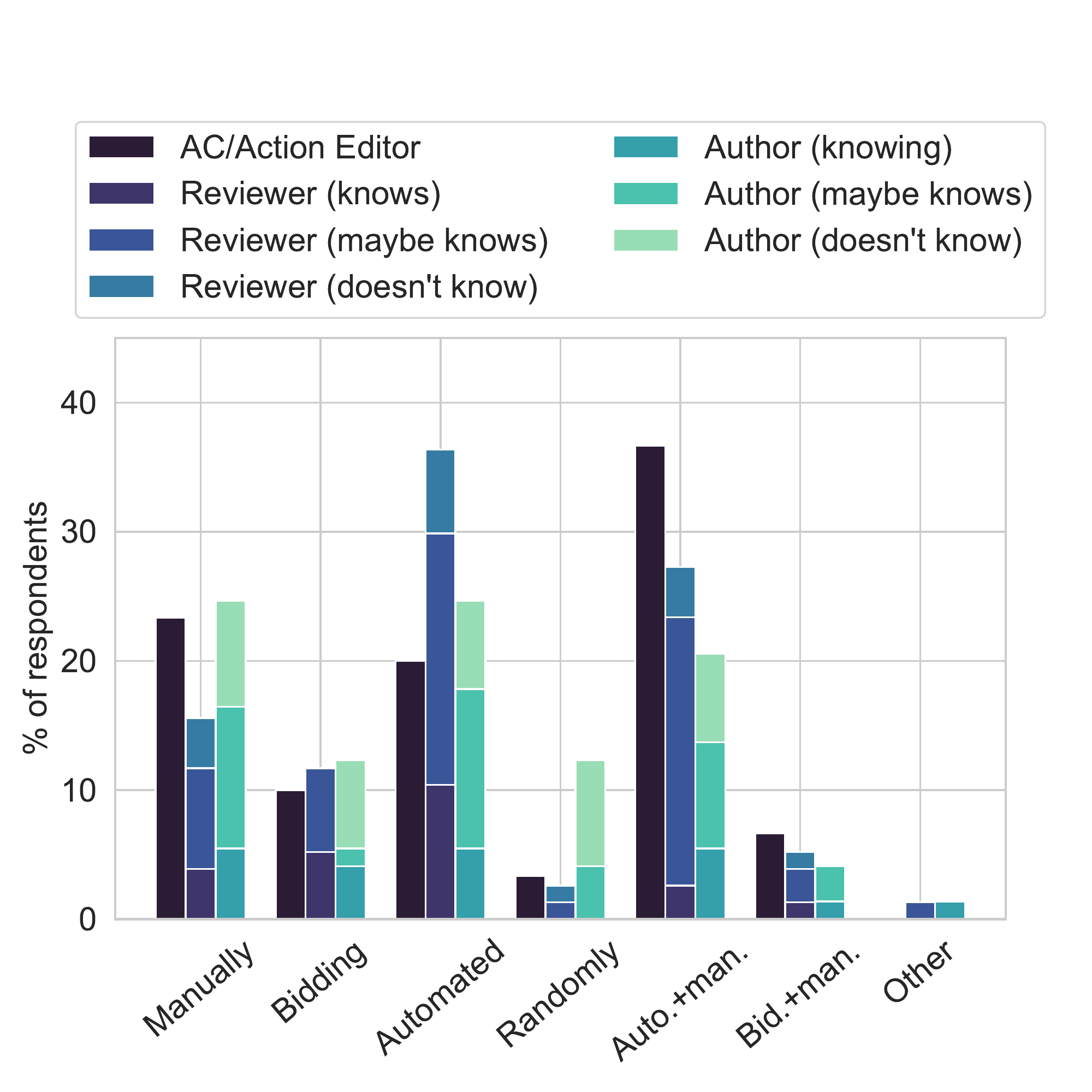}
     \caption{We ask reviewers and authors whether they know for certain, or maybe knows, or do not know how the paper-reviewer matching was done for their last CL/NLP venue. We then ask both reviewers, authors and ACs what they believe (or knows in the case of some) was the process for this venue The guesses, and knowledge herof, are much different from \textit{best} options in \cref{fig:whatshould}, discussed in \cref{sec:recommendations}.
     }
    \label{fig:whatis}         
 \end{figure}
 
 \begin{figure}[h!]
    \centering
    \includegraphics[width=0.9\linewidth]{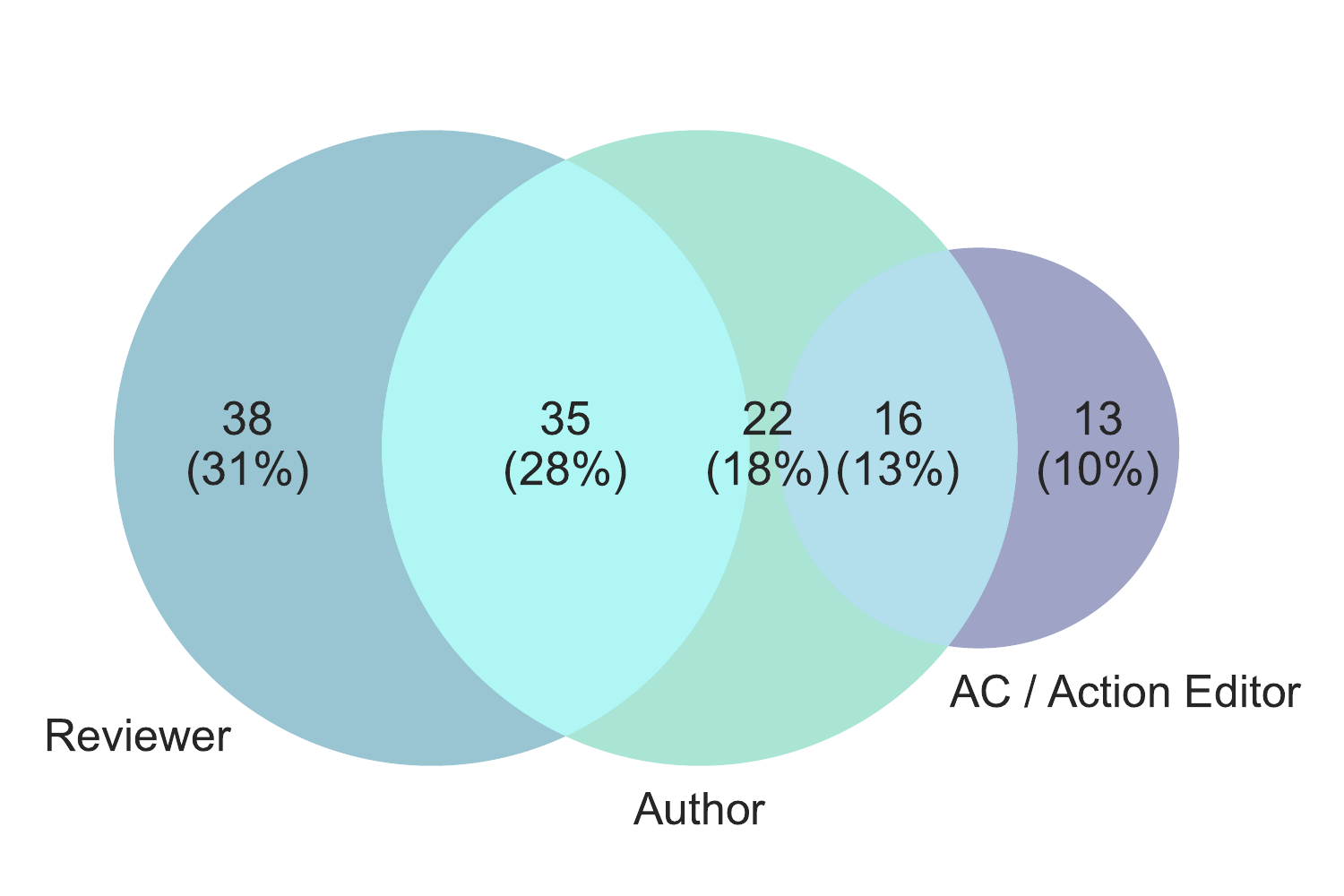}
    \caption{Distribution of \textbf{non-students} in the three roles, with overlap derived from asking the question \textit{Did you also serve as a reviewer/author?} for their last CL/NLP venue.}
    \label{fig:vennSenior}
\end{figure}

\begin{figure}[ht]
    \centering
    \begin{subfigure}{\linewidth}
         \centering
         \includegraphics[width=\linewidth]{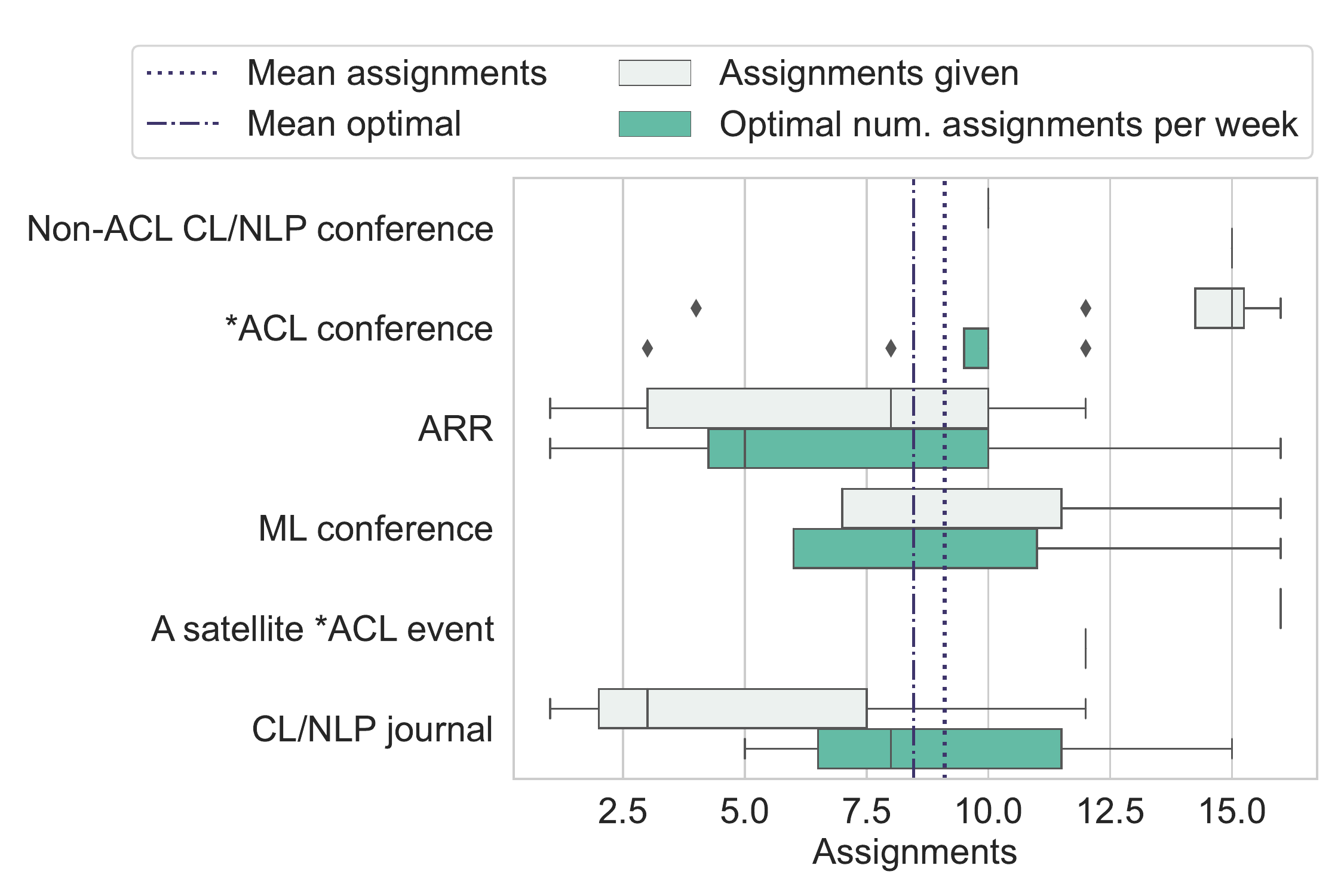}
         \caption{AC/Action Editors}
     \end{subfigure}
    \begin{subfigure}{\linewidth}
         \centering
         \includegraphics[width=\linewidth]{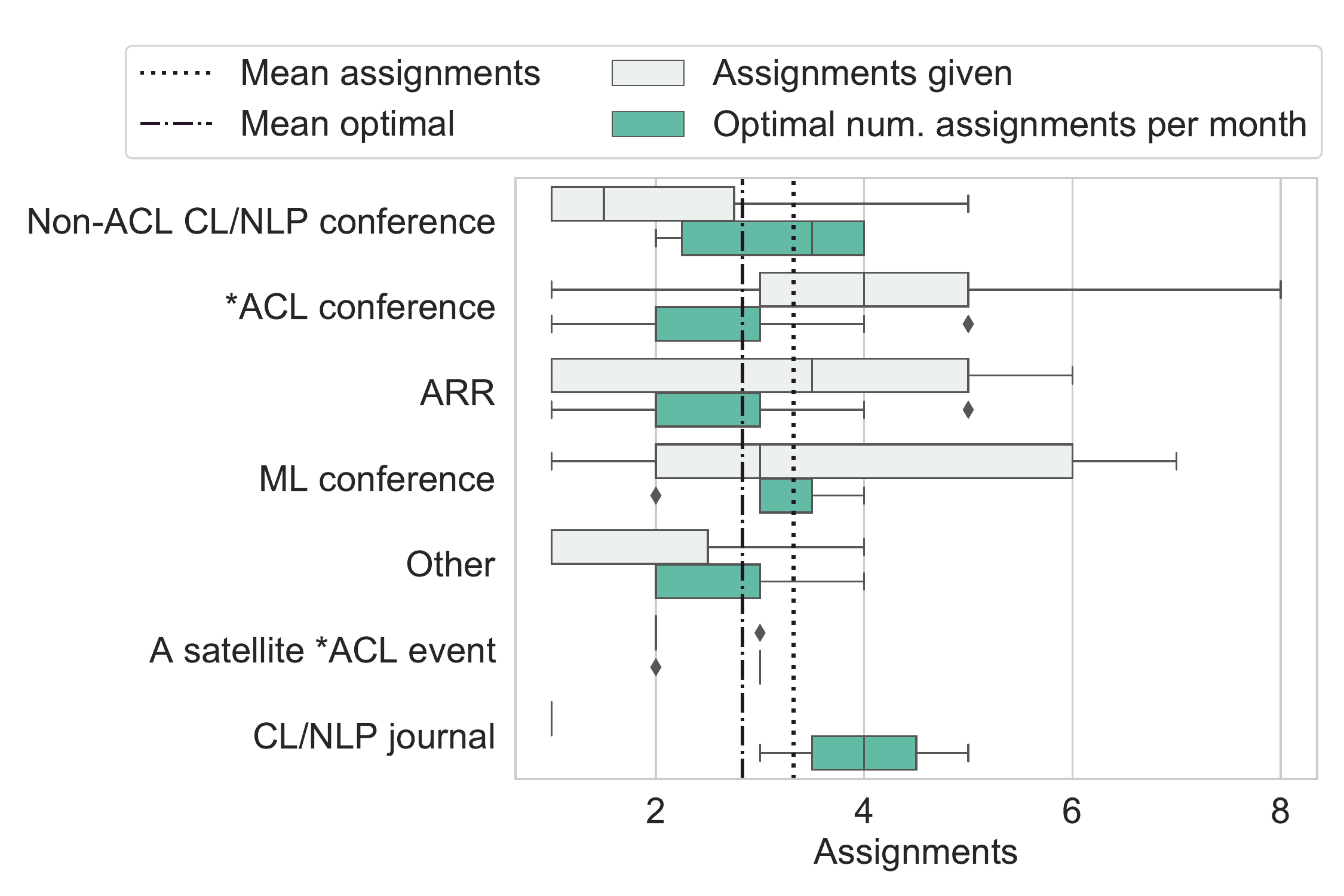}
         \caption{Reviewers}
     \end{subfigure}
    \caption{The boxplots shows the number of assignments given and optimal for a) ACs and b) Reviewers, discussed in \cref{sec:workload}. Number of given assignments are reported for the venue in which the respondent last served as AC/reviewer, and optimal number of assignments are reported for time periods one week for ACs and one month for reviewers. Mean given and optimal number of assignments, across all respondents/venues, are shown with vertical striped lines. 
    }
    \label{fig:revload}
\end{figure}

\begin{figure*}
    \centering
    \begin{subfigure}[b]{\linewidth}
         \centering
         \includegraphics[width=\linewidth]{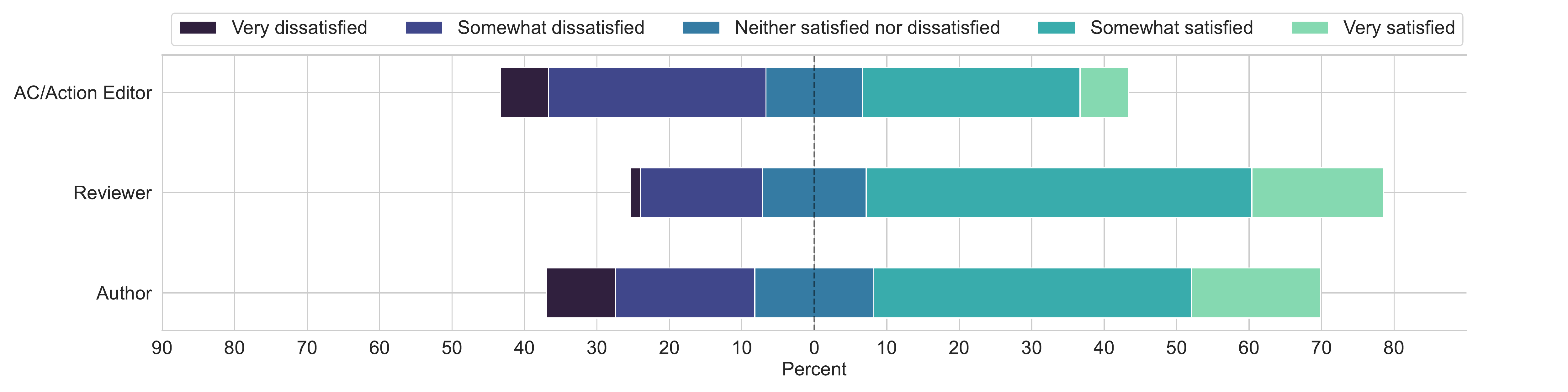}
         \caption{Reported satisfaction with most recent experience. AC / Action Editor question: \textit{(...) how satisfied were you with the support provided to you to improve the paper-reviewer matching?} Reviewer question: \textit{(...) How satisfied were you with the paper-reviewer matching?} Author question: \textit{(...) How satisfied were you with the amount of constructive criticism in the reviews you received?}}
     \end{subfigure}
    \begin{subfigure}[b]{\linewidth}
         \centering
         \includegraphics[width=\linewidth]{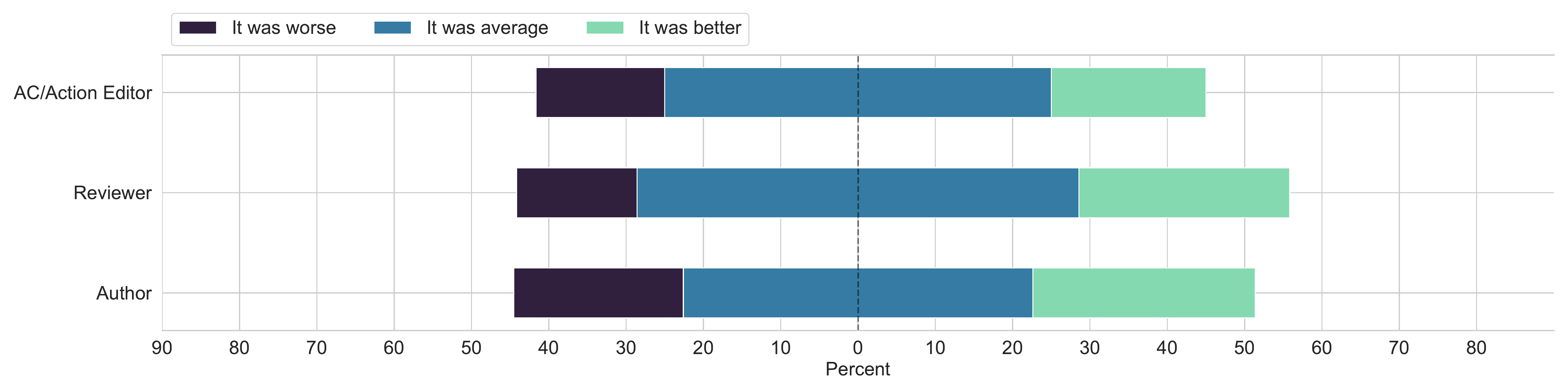}
         \caption{Question: \textit{would you say your most recent experience with paper-reviewer matching/paper assignment(s)/set of reviews described above, was better or worse than on average?}}
     \end{subfigure}
    \caption{We ask all respondents general questions about their satisfaction with their last experience as an AC/reviewer/author and their overall satisfaction in this role. We discuss these results in the limitations \cref{sec:limitation}.}
    \label{fig:satisfaction}
\end{figure*}

\clearpage

\begin{table*}[t]
\begin{scriptsize}
\begin{tabularx}{\textwidth}{l|XXX|XXX|XXX}
\toprule
& \multicolumn{3}{c}{{\sc Area Chair}} & \multicolumn{3}{c}{{\sc Reviewer}} & \multicolumn{3}{c}{{\sc Author}}\\

 & 1-5 &  6-10  & >10 & 1-5 &  6-10 & >10 & 1-5 & 6-10 & >10\\

\midrule
Bachelor's student &0 & 0& 0 &0 &0 &0 &0&1&0\\

Master's student &0&0&0&1&0&0&0&0&0\\

PhD student &0&1&0  & 7&10&3 & 25&6&2\\

Other &0&0&0 & 0&0&1 & 2&0&1\\

Postdoc & 0&1&2 & 1&3&12 & 2&0&11\\

Academic researcher & 0&0&19 & 4&1&21 & 4&0&13\\

Industry researcher & 1&0&6 & 5&0&8 & 1&2&3\\
\midrule 

Total & 1&2&27 & 18&14&45 & 34&9&30\\
\bottomrule
\end{tabularx}
    \caption{This table shows the count of respondents from each role (AC/reviewer/author) reporting one of 7 career statuses and an amount of times having had their own papers reviewed. The numbers reflects those plotted in \cref{fig:career}, \cref{sec:who}.}
    \label{tab:background}
\end{scriptsize}
\end{table*}

\begin{table*}[]
\begin{scriptsize}

    \centering
    \begin{tabular}{l|ccccccc}
\toprule
{} &  Tasks &   Bidding & Method &   Type of paper &   Trust & Data &   Affinity Score  \\
\midrule
AC / Action Editor &  3.90$\pm$0.83 &  3.67$\pm$1.25 &  3.70$\pm$0.74 & 3.37$\pm$0.95 &  3.27$\pm$1.67 &   2.83$\pm$1.00 &  2.13$\pm$1.50 \\
Reviewer &  4.00$\pm$0.90 & 4.03$\pm$1.07 & 3.86$\pm$0.85 & 3.16$\pm$0.97 & 2.96$\pm$1.57 & 2.75$\pm$1.09 & 1.97$\pm$1.38 \\
Author &   3.95$\pm$0.86 & 3.86$\pm$1.22 & 3.59$\pm$0.87 & 3.45$\pm$1.18 & 3.11$\pm$1.60 &   2.84$\pm$0.98 &  1.85$\pm$1.32 \\

\midrule
Grand mean  & 3.95 &3.85 & 3.72 & 3.32 & 3.11 &    2.81 & 1.99 \\
\bottomrule
\end{tabular}
    \caption{Mean importance with 0=Not sure, 1=Not important, 2=Slightly important, 3=Moderately important, 4=Very important and 5=Extremely important, for the seven paper-reviewer matching factors shown in \cref{fig:importances}. Removing "Not sure" does not change the overall ranking. The grand mean is the unweighted mean of ACs', reviewers' and authors' mean scores. The mean absolute difference is greatest between ACs and reviewers (0.20) and smallest between ACs and authors (0.12), while between reviewers and authors it is 0.16. These results are discussed in \cref{sec:ideal} under ``Stakeholder preferences".}
    \label{tab:meanimportancescores}
\end{scriptsize}
\end{table*}

\begin{table*}[]
\begin{scriptsize}
    \centering
    \begin{tabular}{l|ccccccc}
        \toprule
        {} &  Tasks &  Bidding &  Method &  Type of Paper &  Trust &   Data &  Affinity Score \\
        \midrule

        Computationally-aided linguistic analysis &  1.000 &    0.624 &    0.205 &       0.089 &  0.699 &  \textbf{0.035$^<$} &       0.654 \\
        \midrule[0.1pt]

        NLP engineering experiment paper          &  1.000 &    0.716 &    \textbf{0.038$^>$} &       0.135 &  0.270 &  0.772 &       0.699 \\
        \midrule[0.1pt]

        Reproduction paper                        &  0.457 &    0.766 &    0.055 &       0.601 &  1.000 &  \textbf{0.004$^>$} &       0.562 \\
        \midrule[0.1pt]

        Resource paper                            &  0.728 &    0.433 &    0.727 &       0.162 &  0.818 &  1.000 &       0.616 \\
        \midrule[0.1pt]

        Position paper                            &  0.222 &    0.766 &    0.433 &       0.135 &  0.236 &  0.493 &      \textbf{0.003$^<$} \\
        \midrule[0.1pt]

        Survey paper                              &  1.000 &    0.186 &    0.738 &       0.420 &  1.000 &  0.840 &       0.480 \\
        \midrule[0.1pt]

        Other                                     &  0.601 &    0.052 &    1.000 &       \textbf{0.019$^<$} &  0.733 &  0.202 &       0.063 \\
        \bottomrule
        \end{tabular}
    \caption{P-values of two-sided Fisher Exact tests, discussed in \cref{sec:ideal}. For each contribution type, we test the null hypothesis that there is no difference in whether respondents find a paper-match factor (from \cref{fig:importances}) more than or less than moderately important, depending on whether or not individuals have worked on the specific types of papers (contribution types). For each contribution type $i$ and paper-match factor $j$, a $2\times2$ contingency table is made with the counts of a) respondents having worked with type $i$ and finding factor $j$ \textit{less} than moderately important, b) having worked with type $i$ and finding factor $j$ \textit{more} than moderately important, c) having \textit{not} worked with $i$ and finding $j$ \textit{less} than moderately important, d) having \textit{not} worked with $i$ and finding $j$ \textit{more} than moderately important. The p-values reflect the probability of observing the given counts or something more imbalanced between those having and not having worked on type $i$. %
    Significant p-values, $p<0.05$, are in bold, and for these, superscript $>$ denotes that respondents having worked with $i$ believe factor $j$ is \textit{more} than moderately important, and the superscript $<$ denotes the opposite.}
    \label{tab:fisher1}
    
\end{scriptsize}
\end{table*}

\end{document}